\documentclass[10pt,twocolumn,letterpaper]{article}

\usepackage{cvpr}              
\usepackage{stfloats}

\usepackage{makecell}
\usepackage{cuted}

\definecolor{cvprblue}{rgb}{0.21,0.49,0.74}
\usepackage[pagebackref,breaklinks,colorlinks,allcolors=cvprblue]{hyperref}

\def\paperID{} 
\def\confName{CVPR}
\def\confYear{2026}

\title{PointGS: Semantic-Consistent Unsupervised 3D Point Cloud Segmentation
with 3D Gaussian Splatting }

\author{
Yixiao Song$^1$, \ 
Qingyong Li$^{1,2}$ ,\ \ 
Wen Wang$^{1,}$\thanks{Corresponding authors: Wen Wang (wangwen@bjtu.edu.cn) and Qingyong Li (liqy@bjtu.edu.cn).\\
\hspace*{1.5em}\textsuperscript{1}The code is available at: \url{https://github.com/SebastianYIXIAO/pointGS}}\ ,\ \ 
Zhicheng Yan$^1$\\
$^1$Key Laboratory of Big Data \& Artificial Intelligence in Transportation \\(Beijing Jiaotong University), Ministry of Education\\
$^2$Frontiers Science Center for Smart High-speed Railway System, Beijing Jiaotong University\\
{\tt\small (songyixiao, liqy, wangwen, yanzhicheng)@bjtu.edu.cn}
}

\begin{document}
\maketitle
\begin{abstract}

Unsupervised point cloud segmentation is critical for embodied artificial intelligence and autonomous driving, as it mitigates the prohibitive cost of dense point-level annotations required by fully supervised methods. While integrating 2D pre-trained models such as the Segment Anything Model (SAM) to supplement semantic information is a natural choice, yet this approach faces a fundamental mismatch between discrete 3D points and continuous 2D images. This mismatch leads to inevitable projection overlap and complex modality alignment, resulting in compromised semantic consistency across 2D-3D transfer. To address these limitations, this paper proposes \textbf{PointGS}, a simple yet effective pipeline for unsupervised 3D point cloud segmentation. PointGS leverages 3D Gaussian Splatting as a unified intermediate representation to bridge the discrete-continuous domain gap. Input sparse point clouds are first reconstructed into dense 3D Gaussian spaces via multi-view observations, filling spatial gaps and encoding occlusion relationships to eliminate projection-induced semantic conflation. Multi-view dense images are rendered from the Gaussian space, with 2D semantic masks extracted via SAM, and semantics are distilled to 3D Gaussian primitives through contrastive learning to ensure consistent semantic assignments across different views. The Gaussian space is aligned with the original point cloud via two-step registration, and point semantics are assigned through nearest-neighbor search on labeled Gaussians. Experiments demonstrate that PointGS outperforms state-of-the-art unsupervised methods, achieving +0.9\% mIoU on ScanNet-V2 and +2.8\% mIoU on S3DIS.
\end{abstract}
  
\section{Introduction}
\label{sec:intro}

\begin{figure}[t]
    \vspace*{-0in} 
    \hfill 
    \makebox[0pt][r]{\includegraphics[width=\linewidth]{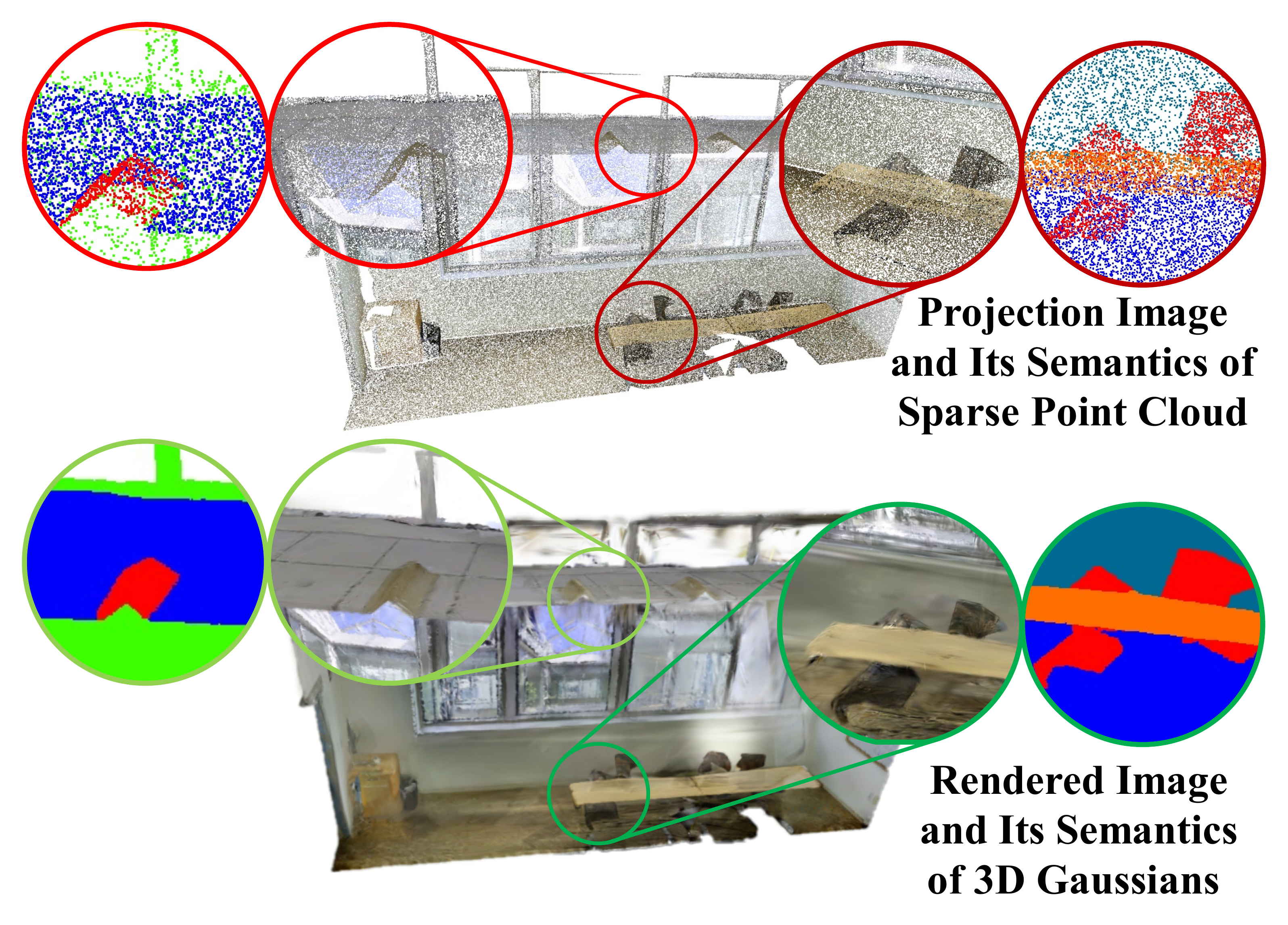}} 
    \caption{In the conference room scene, the upper part of the figure shows that the sparse point cloud causes the foreground points and background points to overlap, while the lower part shows that in the Gaussian space, the background is completely blocked by the foreground.}
    \label{fig:motivation}
\end{figure}

Point cloud segmentation involves classifying individual points and is a critical task in emerging fields such as embodied intelligence and autonomous driving. Fully supervised point cloud segmentation methods have evolved significantly over the years, from pioneering works ~\cite{qi2017pointnet, qi2017pointnet++} to more recent advancements leveraging multimodal fusion~\cite{tang2023prototransfer,he2024segpoint,wu2024dino,xia2024gsva} and Transformer architectures~\cite{park2023self,yu2022pointbert,wu2024pointv3}. The current fully-supervised methods enable a finer-grained understanding of complex 3D structures. However, these methods require dense point-wise annotations, which are time-consuming and computationally expensive to obtain, especially for large-scale scenes.

To alleviate the burden of annotating a large number of samples, some studies have shifted towards unsupervised methods, which mainly rely on methods of local feature extraction and clustering~\cite{liu2023u3ds3, zhang2023growsp}. Without access to high-level semantic priors, these methods only capture local geometric patterns and prioritize geometric similarity in clustering. 
As a result, they struggle to distinguish objects that are geometrically similar but semantically distinct.
For instance, \emph{wall} and \emph{board}, which are both planar yet functionally and semantically distinct, remain indistinguishable. 
Fortunately, the 2D vision domain has accumulated massive labeled data and developed generalizable pre-trained large models (e.g., DINOv2~\cite {oquab2023dinov2}, SAM~\cite {kirillov2023segment}), which can provide rich semantic information to compensate for this shortcoming. Recent works have attempted to integrate 2D modal pretrained models into unsupervised point cloud segmentation~\cite{yu2022p2p, rozenberszki2024unscene3d, yin2024sai3d, zhang2025logosp}.

Nevertheless, existing cross-modal methods face a fundamental mismatch between sparse 3D point clouds and continuous 2D images. Sparse points capture limited geometry, leaving large gaps that hinder encoding occlusion and depth order, causing points from different semantic categories to overlap in 2D projections (Fig. \ref{fig:motivation}, top). SAM, trained on dense 2D data, generates masks directly from the semantically ambiguous projection images, which leads to poor 3D segmentation results. Moreover, the inherent domain gap between discrete 3D points and continuous 2D pixels demands complex alignment or additional 3D pre-training to resolve ambiguous point-pixel correspondences, increasing complexity and compromising semantic consistency.

To address this core mismatch, we draw inspiration from 3D Gaussian Splatting (3D-GS)~\cite{kerbl20233dgaussians}, an advanced 3D reconstruction technique. Its dual core properties directly resolve the limitation above. First, 3D-GS replaces discrete point clouds with dense 3D Gaussian ellipsoid primitives that provide continuous coverage in local space. These primitives fill spatial gaps and encode occlusion, enabling dense, occlusion-aware 2D renderings (Fig. \ref{fig:motivation}, bottom), where foreground primitives block background signals and prevent mixed-semantic masks from SAM. Second, 3D-GS supports differentiable rendering, preserving native 3D spatial relationships in 2D images so that distilled semantics inherit 3D consistency. These two properties together bridge the discrete-continuous domain gap, eliminating the need for complex 2D-3D alignment or extra 3D pre-training.

Based on the analysis above, this paper introduces a simple yet effective pipeline \textbf{PointGS} that achieves semantic-consistent unsupervised 3D point cloud segmentation with 3D Gaussian Splatting. PointGS follows three key steps to reduce the reliance of unsupervised point cloud segmentation on local geometric feature learning. The input sparse point cloud is first reconstructed into a dense 3D Gaussian space using multi-view observations. This step leverages 3D-GS’s fast training speed and continuous scene representation to fill spatial gaps in raw point clouds. Then, multi-view dense images are rendered from the Gaussian space and SAM is employed to extract 2D semantic masks. Contrastive learning is applied to distill these 2D semantics to 3D Gaussian primitives, ensuring consistent semantic assignments across different views. Finally, we extract center points of 3D Gaussian ellipsoids and perform a two-step registration to align the Gaussian space with the original point cloud. Semantics of each point are assigned via nearest-neighbor search on the semantically labeled Gaussians. Extensive experiments on widely-used indoor databases, S3DIS and ScanNet-v2, validate the effectiveness of PointGS. It outperforms state-of-the-art unsupervised methods by +2.8\% mIoU on S3DIS and +0.9\% mIoU on ScanNet-v2. 

In summary, our contributions are as follows: 
\begin{itemize}
\item We leverage Gaussian Splatting as a unified intermediate representation for unsupervised point cloud segmentation, effectively bridging the discrete-continuous domain gap between 3D points and 2D images.
\item We propose the PointGS framework that integrates 3D-GS with multi-view SAM segmentation. It realizes semantic distillation and accurate Gaussian-point alignment without complex preprocessing or manual intervention.
\item On two widely-used benchmark datasets, this method achieves significant improvements compared to existing unsupervised methods in 3D point cloud semantic segmentation, demonstrating its effectiveness.
\end{itemize}

\section{Related Work}
\label{sec: Related Work}

\textbf{Supervised 3D Point Cloud Semantic Segmentation.} Supervised 3D point cloud semantic segmentation has evolved from early point-based methods \cite{qi2017pointnet, qi2017pointnet++} using MLPs and max-pooling, extended by hierarchical grouping \cite{qi2017pointnet++}, to deformable/sparse convolutions \cite{thomas2019kpconv, hu2020randla, peng2024oa} for efficient large-scale processing. Recent transformer-based architectures \cite{zhao2021pt, wu2022ptv2, yu2022pointbert, wu2024pointv3} improve point cloud understanding through attention-based modeling, while Stratified Transformer \cite{lai2022stratified} and PointNeXt \cite{qian2022pointnext} further enhance feature extraction and efficiency via long-range context modeling and effective model scaling, respectively. Mamba3D \cite{han2024mamba3d} integrates the Mamba State Space Model into 3D point cloud processing and replaces the traditional Transformer's quadratic attention calculation with a global receptive field and linear complexity. Surveys \cite{sarker2024comprehensive} highlight these advances, but all require extensive annotations on datasets like ScanNet \cite{dai2017scannet} and S3DIS \cite{armeni20163d}. However, supervised methods demand extensive labeled data, which is expensive and labor-intensive to obtain for 3D point clouds due to their high dimensionality and variability. This limitation hinders generalization to new domains, prompting the shift to unsupervised approaches that exploit inherent data structures for label-free segmentation.

\smallskip
\noindent\textbf{Unsupervised Methods Based on Superpoints and Clustering.} Unsupervised segmentation often begins with clustering techniques that group points into superpoints for semantic discovery. GrowSP \cite{zhang2023growsp} employs iterative superpoint merging based on feature similarity, achieving robust results on indoor datasets. U3DS$^3$ \cite{liu2023u3ds3} leverages dual invariant-equivariant paths for iterative refinement. LogoSP \cite{zhang2025logosp} uses local-global superpoint grouping in the frequency domain for pseudo-label generation. These methods, as reviewed in \cite{wang2023survey, zhang2024advancements}, are scalable but sensitive to initial groupings and category imbalances.

\smallskip
\noindent\textbf{Unsupervised Methods with 2D Prior.} 
Image-based methods link 3D points and 2D images, leveraging 2D information without 3D pretraining, offering efficiency advantages. P2P \cite{yu2022p2p} converts 3D point clouds into color images that retain geometric information and uses pre-trained 2D image models for end-to-end optimization. Thus, it can transfer 2D knowledge to 3D point cloud analysis tasks with a lower parameter cost. PointDC \cite{chen2023pointdc} distills multi-view 2D features into 3D point representations via cross-modal distillation. UnScene3D \cite{rozenberszki2024unscene3d} uses self-supervised color and geometry features for instance discovery. CluRender \cite{mei2024unsupervised} combines clustering with multi-view rendering. Segment3D \cite{huang2024segment3d} projects the masks generated by SAM onto the point cloud to obtain 3D pseudo-labels for pre-training . Reviews \cite{guo2023unsupervised, zhang2025point} praise semantic richness but identify issues like occlusion causing sparse overlaps and semantic confusion in back-interpolation. Directly adopting the 2D models such as SAM \cite{kirillov2023segment} may lead to view-inconsistent labels and weakened spatial consistency.

\section{Method}

\begin{figure*}[t]
  \centering
  \includegraphics[width=1\linewidth]{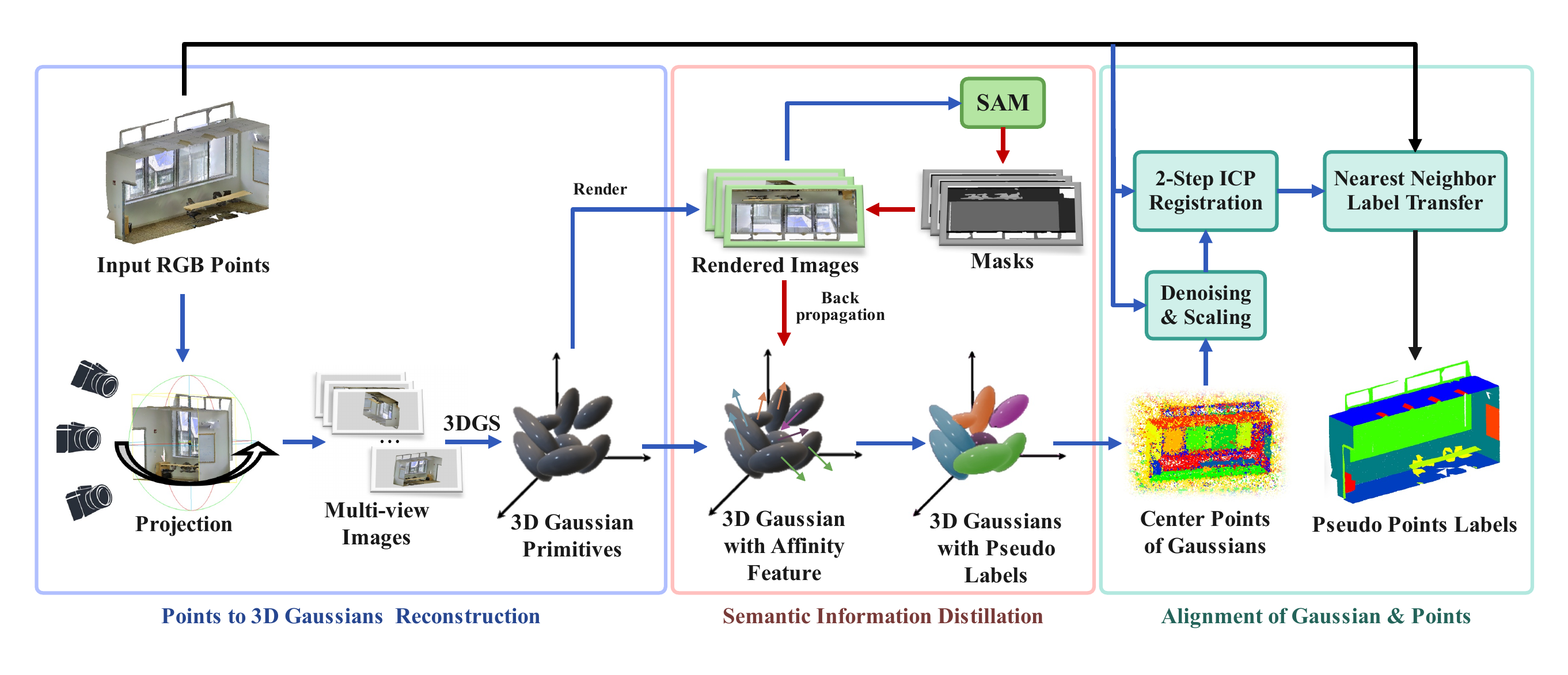}
  \caption{
    \textbf{The pipeline of our method.}
    Given an indoor point cloud, we first generate multi-view projections and apply Gaussian splatting.
    Then, the rendered images are segmented by SAM, and semantic cues are transferred back to 3D Gaussians.
    Finally, the segmented Gaussians are refined and aligned with the raw point cloud through the alignment module of Gaussians and points, and labels are propagated back to the original points.
  }
  \label{fig: pipeline}
\end{figure*}

In this work, we study the problem of unsupervised semantic segmentation of indoor point clouds, where no human labels and point cloud datasets for pre-training are available. Our approach combines 2D segmentation priors with 3D Gaussian splatting to address the shortcomings of current 2D prior-guided point cloud segmentation methods. Specifically, we first introduce the problem definition in Sec.~3.1. We then revisit the Gaussian splatting formulation and rendering process, which provides a differentiable 3D representation suitable for semantic transfer 
(Sec.~3.3). To distill semantic information from SAM, we utilize segmentation masks predicted from multi-view 2D rendered images to supervise the alignment of affinity features of 3D Gaussians that belong to the same semantics. (Sec.~3.4). Next, we extract the geometric center points of the 3D Gaussian that contain semantic information, generate the point cloud distribution of Gaussians, and align the Gaussian point cloud with the original point cloud through a two-stage ICP registration. Finally, the labels in the Gaussian point cloud are assigned to the original point cloud through nearest neighbor calculation (Sec.~3.5). An overview of the proposed framework is illustrated in Fig.~\ref{fig: pipeline}.

\subsection{Problem Statement}
Let the raw indoor point cloud be defined as
\begin{equation}
\mathcal{P} = \{ \mathbf{p}_i \}_{i=1}^N, \quad \mathbf{p}_i \in \mathbb{R}^6,
\label{eq:}
\end{equation}
where $N$ is the number of points, and each point contains XYZ coordinates with RGB values. The goal of unsupervised semantic segmentation is to assign each point $\mathbf{p}_i$ a semantic label $l_i \in \{1, \dots, K\}$ without manual supervision, where $K$ is the number of semantic categories.

\subsection{Preliminary}
\noindent\textbf{3D Gaussian Splatting (3D-GS).} 
3D Gaussian Splatting (3D-GS) reconstructs dense 3D scenes from multi-view 2D images and camera parameters \cite{kerbl20233dgaussians}. It models scenes via $N$ number of 3D colored Gaussians $\mathcal{G} = \{g_1, g_2, \dots, g_N\}$, where the position of each Gaussian is specified by its mean, and its shape is described by the covariance. A key advantage is differentiable rasterization, which projects 3D Gaussians to 2D image planes and computes pixel colors via alpha compositing (depth-sorted blending of Gaussian colors and opacities). For a given pixel $u$, let $\mathcal{G}_u$ denote the set of Gaussians influencing $u$. The color $C(u)$ is calculated as follows:
\begin{equation}
C(u) = \sum_{i=1}^{|\mathcal{G}_u|} \alpha_{g_i}(u)\,\mathbf{c}_{g_i}(u) \prod_{j=1}^{i-1}\big(1-\alpha_{g_j}(u)\big),
\label{eq:}
\end{equation}
where $g^u_i$ is the $i$-th Gaussian in $\mathcal{G}_u$, $\mathbf{c}_{g_i}(u)$ is the view-dependent color of $g^u_i$, and $\alpha_{g_i}(u)$ is the trainable opacity (derived from projected covariance), and the product term accounts for foreground transparency. Critical to our work, 3D-GS rendering is differentiable and supports back-propagation: differentiability enables gradient propagation from 2D pixels to 3D Gaussians, while the explicit Gaussian representation enables bidirectional association between 2D SAM masks and 3D Gaussians. Additionally, 3D-GS generates denser, more continuous scene representations than direct sparse point cloud projections, ensuring reliable 2D-to-3D semantic transfer.

\smallskip
\noindent\textbf{Scale-Conditioned 3D Gaussian Affinity Features.}
To handle multi-granularity ambiguity in lifting 2D segmentation priors to 3D Gaussians—where a single Gaussian may belong to different objects or parts depending on the scale, GARField~\cite{kim2024garfield} introduces scale-conditioned feature fields. It computes the scale $s_M$ of a 2D mask $M$ in a view-consistent manner by projecting $M$ into 3D space using camera intrinsics and depth estimates from a pre-trained radiance field, yielding a point cloud $P$. The scale is then defined as:
\begin{equation}
s_M = 2 \sqrt{\mathrm{std}(X(P))^2 + \mathrm{std}(Y(P))^2 + \mathrm{std}(Z(P))^2},
\label{eq:}
\end{equation}
where $X(P)$, $Y(P)$, and $Z(P)$ are the coordinate components of $P$, and $\mathrm{std}(\cdot)$ is the standard deviation. This allows features to adapt to varying granularities by conditioning on $s_M$, but GARField's reliance on implicit fields requires repeated queries for different scales, limiting efficiency.

Building upon this foundation, SAGA~\cite{cen2025saga} operationalizes scale conditioning for 3D-GS, primarily for prompt-guided 3D segmentation tasks where user inputs guide mask generation.
It attaches learnable affinity features $\mathbf{f}_g \in \mathbb{R}^D$ to each explicit Gaussian $g$ in 3D-GS, where $D$ is the feature dimension. To resolve multi-granularity ambiguity, a soft scale-gate $S: [0,1] \to [0,1]^D$ modulates features. Implemented as a linear layer followed by a sigmoid, the gate produces $\mathbf{f}_g^s = S(s) \odot \mathbf{f}_g$, where $\odot$ denotes the Hadamard product. 

These features are optimized through scale-aware contrastive learning. 3D features are rendered to 2D pixels 
$u$ as $F(u) = \sum_i \mathbf{f}_{g_i} \alpha_{g_i} \prod_{j<i} (1 - \alpha_{g_j})$, then gated to $F^s(u) = S(s) \odot F(u)$. Supervision uses correspondences from scale-sorted masks, with the loss:
\begin{multline}
\mathcal{L}_{\mathrm{corr}}(s, u_1, u_2) = (1 - 2 \cdot \mathrm{Corr}_m(s, u_1, u_2))\\
\cdot \max(\mathrm{Corr}_f(s, u_1, u_2), 0),
\label{eq:saga_loss}
\end{multline}
where $\mathrm{Corr}_m$ is 1 if pixels share a mask (derived from identity vectors) and 0 otherwise, and $\mathrm{Corr}_f$ is the cosine similarity of gated features. A norm regularization $\mathcal{L}_{\mathrm{norm}}(u) = 1 - \|F(u)\|_2$, applied during rendering with normalized $\mathbf{f}_g$, promotes alignment between 2D and 3D spaces.

This scale-conditioned affinity mechanism directly links 3D Gaussian representations to scale-aware semantic cues, offering a robust solution to multi-granularity ambiguity that is well-suited for 2D-to-3D semantic transfer. For our task, we leverage this mechanism but adapt it to our unsupervised scenario. We use automatically generated SAM masks without user prompts and focus on transferring the resulting semantic information to raw point clouds.

\subsection{Points to 3D Gaussians Reconstruction}
We first project the point cloud according to a predefined sequence of viewing angles to obtain multi-view images. These images are then used for scene reconstruction via 3D-GS. After obtaining the 3D Gaussian scene, we render images from corresponding viewpoints. Since the rendering process is differentiable and supports back-propagation, the rendered images contain the spatial information of the 3D Gaussian scene. Moreover, compared to images obtained by direct projection of sparse point clouds, the rendered images from 3D-GS exhibit more continuous and dense object semantics. In addition, we introduce a Multi-View Consistency Check inspired by SuGaR~\cite{guedon2024sugar}. We screen the 3D Gaussians using adjacent multiple rendered views. If there are more than three adjacent views where the Gaussian primitives are not involved in the rendering process, then they will be deleted. This method can eliminate some Gaussian noise and reduce the influence of the background on the foreground when migrating 2D semantics to 3D space.

\subsection{Semantic Information Distillation}
To inject semantic priors into the 3D Gaussian space, we employ the Segment Anything Model (SAM) to segment the rendered 2D projections. Let $\mathcal{M}^{(v)} = \{ M_j^{(v)} \}$ denote the set of semantic masks predicted by SAM from the $v$-th view, where each $M_j^{(v)} \in \{0,1\}^{H \times W}$ is a binary mask with height $H$ and width $W$. These 2D masks provide rich semantic information but are view-specific and must be backpropagated to the 3D Gaussians. We achieve this through a scale-aware contrastive learning framework, building on the distillation strategy in SAGA~\cite{cen2025saga}. Specifically, we attach a learnable affinity feature $\mathbf{f}_g \in \mathbb{R}^D$ to each Gaussian $g$. To handle multi-granularity ambiguity, a scale gate $S(s)$ is implemented as a linear layer, which modulates the affinity features as $\mathbf{f}_g^s$ for a given scale $s$. For supervision, we compute pixel correspondences from sorted masks based on their 3D scales $s_{M_j^{(v)}}$. The mask correspondence between pixels $u_1$ and $u_2$ at scale $s$ is $\text{Corr}_m(s, u_1, u_2) = 1$ if they share at least one mask, and $0$ otherwise. The feature correspondence is the cosine similarity:
\begin{equation}
\text{Corr}_f(s, u_1, u_2) = \langle F^s(u_1), F^s(u_2) \rangle. 
\end{equation}
The contrastive loss is the same as that in Eq. ~(\ref{eq:saga_loss}). The total loss is summed over sampled pixel pairs and pixels in each view with regularization on the rendered feature norm. This process distills semantic information from SAM to 3D Gaussians and obtains 3D Gaussians with pseudo-labels. In addition, several concurrent works follow a similar 2D to 3D mask lifting paradigm, including Gaussian Grouping~\cite{ye2024gaussiangrouping}, FlashSplat~\cite{shen2024flashsplat}, and COB-GS~\cite{zhang2025cob}.

\subsection{Gaussian-to-Point Cloud Alignment}
Next, the semantic information of Gaussians needs to be transferred to the original point cloud. But the 3D Gaussian coordinate system is not directly aligned with that of the original point cloud. In particular, their spatial scale and orientation may differ due to the rendering and reconstruction process. To effectively transfer semantic information from the 3D Gaussians to the raw point cloud, we design an alignment procedure that first extracts the geometric center points $P_{\mathcal{G}}$ of 3D Gaussian primitives and then removes noisy points and adjusts their scale. Finally, we align the rescaled Gaussian points to the raw point cloud through the 2-Step ICP registration.

\smallskip
\noindent\textbf{Gaussian Center Points Density Denoising and Scaling.} 
To eliminate the interference caused by Gaussian noise points, we first apply density estimation by calculating the distance between points: 
\begin{equation}
\hat{\rho}_i = \sum_{j \neq i} \exp\left(-\frac{\|\mathbf{p}_i - \mathbf{p}_j\|_2^2}{2h^2}\right).
\end{equation}
where $h$ represents the bandwidth parameter for density estimation. Then, based on the high-density distribution characteristic of the 3D Gaussian distribution at the scene edges, we remove noisy Gaussian points while preserving the contour edges of the scene structure:
\begin{equation}
P_{\mathcal{G}}' = \{\mathbf{p}_i \in P_{\mathcal{G}} \mid \hat{\rho}_i \geq \tau \}.
\label{eq:7}
\end{equation}
where $\tau$ is the preset density threshold. Next, we rescale the Gaussian points $P_{\mathcal{G}}$ by calculating the ratio $s$ between the contour edge points and the original point cloud $P_{\mathcal{O}}$:
\begin{equation}
s = \frac{\operatorname{diam}(P_{\mathcal{O}})}{\operatorname{diam}(P_{\mathcal{G}}')},
\quad
P_{\mathcal{G}}^s = \{\bar{\mathbf{p}}_{\mathcal{G}} + s(\mathbf{p}-\bar{\mathbf{p}}_{\mathcal{G}}) \mid \mathbf{p}\in P_{\mathcal{G}} \}.
\label{eq:8}
\end{equation}
\smallskip
\noindent\textbf{Two-Stage ICP Registration.} 
The Gaussian points of the same scale as the raw point cloud and the Gaussian point cloud still require further registration. We first perform a coarse ICP registration~\cite{besl1992icp}, the rotation matrix and the translation vector are: 
\begin{equation}
(R^{(1)}, \mathbf{t}^{(1)}) = \arg \min_{R,\mathbf{t}} \sum_{\mathbf{p} \in P_{\mathcal{G}}^s} \| R\mathbf{p} + \mathbf{t} - \operatorname{NN}(R\mathbf{p}+\mathbf{t}, P_{\mathcal{O}})\|_2^2.
\label{eq:9}
\end{equation}
where $\operatorname{NN}(\cdot, \cdot)$ denotes a function that returns the nearest neighbor of a query point within a target point set. Due to the cubic pattern distribution of indoor scene points, traditional single iteration ICP may get stuck in a local optimum, resulting in incomplete registration. Therefore, we have define a set of six axial directions $\mathcal{D} = \{\pm \mathbf{e}_x, \pm \mathbf{e}_y, \pm \mathbf{e}_z\}$, indexed by k(where k= 1, ..., 6). For each direction, we further apply four rotations with angles $\theta_t \in \{0^\circ,90^\circ,180^\circ,270^\circ\}$ indexed by t(where t= 1, ..., 4). For each of the 24 possible (k, t) combinations, ICP is repeated, and the RMSE is measured:
\begin{equation}
d_i = \| R^{(1)}\mathbf{p}_i + \mathbf{t}^{(1)} - \operatorname{NN}(R^{(1)}\mathbf{p}_i + \mathbf{t}^{(1)}, P_{\mathcal{O}})\|_2,
\label{eq:10}
\end{equation}
The optimal registration result $P_{\mathcal{G}}^{R}$ with $(R^*, \mathbf{t}^*)$ is selected by finding the minimum RMSE:
\begin{equation}
E_{k,j} = \sqrt{\frac{1}{|P_{\mathcal{G}}^{(k,t,0)}|} \sum_{\mathbf{p}_i \in P_{\mathcal{G}}^{(k,t,0)}} d_i^2 }.
\label{eq:11}
\end{equation}

\smallskip
\noindent\textbf{Nearest Neighbor Label Transfer.} 
After aligning the Gaussian centers with the original point cloud, we propagate the semantic labels from the aligned Gaussian centers to the original points via nearest-neighbor assignment. Let $P_G=\{(p_n, l_n^G)\}_{n=1}^{N}$ denote the aligned Gaussian centers with labels, and let $P_O=\{b_m\}_{m=1}^{M}$ denote the original point cloud. For each original point $b_m$, we first find its nearest Gaussian center:
\begin{equation}
n^{*}(m)=\arg\min_{1\le n\le N}\left\| b_m - p_n \right\|_2 ,
\label{eq:nn_match}
\end{equation}
and then assign the corresponding label to that point:
\begin{equation}
l_m^O = l_{n^{*}(m)}^G,\qquad m=1,2,\dots,M,
\label{eq:label_transfer}
\end{equation}
where $l_n^G$ and $l_m^O$ denote the semantic labels of the $n$-th Gaussian center and the $m$-th original point, respectively.

\section{Experiments}

\begin{figure*}[t]
  \centering
  \includegraphics[width=0.95\linewidth]{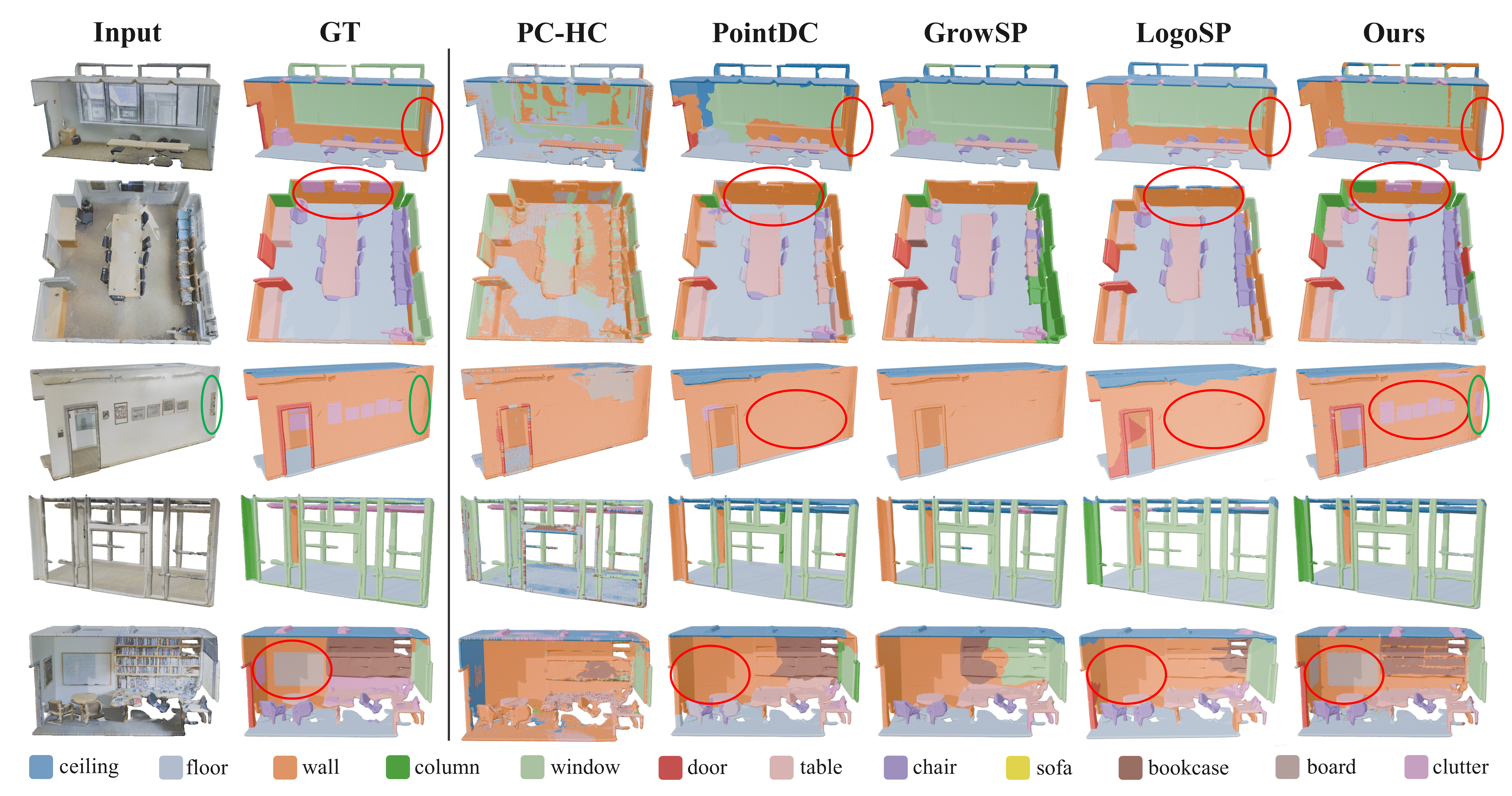}
  \caption{
     Qualitative comparison of unsupervised segmentation on the S3DIS validation set. Each color represents one semantic class. For better understanding, we show some of the color and name matches at the bottom.
  }
  \label{fig:experiment}
\end{figure*}

\subsection{Experiment Details}

\textbf{Dataset and Implementation Details:}
We evaluate our method on two large-scale indoor datasets: ScanNet-v2~\cite{dai2017scannet} and S3DIS~\cite{armeni2017s3dis}. ScanNet-v2 contains 1,613 3D scans from 707 indoor scenes with 20 classes. Following the unsupervised setup, ground-truth labels are not used, and evaluation is performed on the validation set. S3DIS contains 271 scenes with 13 classes. We mainly evaluate on Area 5. To test the generality of our method, we do not use the official multi-view images from either dataset and rely solely on RGB information from point clouds (770×770-pixel projections). Efficiency is measured on a single NVIDIA RTX 3090 GPU. The vanilla 3D-GS performs 43.27 iterations per second, and SAM processes images at 0.35 frames per second. To balance efficiency and quality, we run 3D-GS for 10,000 iterations per scene.

\smallskip
\noindent\textbf{Label Matching and Metric:} As our approach operates in an unsupervised manner, without prior knowledge of the ground truth labels, the resulting clusters may exhibit arbitrary permutations relative to the annotated categories. To evaluate the performance of the unsupervised method, we establish evaluation practices from prior studies. Specifically, for unsupervised clustering, we employ the Hungarian algorithm to optimally align the predicted unlabeled segments with the ground truth labels during evaluation. This alignment enables a robust measurement of semantic consistency between the inferred partitions and the reference annotations, while mitigating the impact of label permutations in the predictions. For evaluation metrics, we report the standard mean Intersection-over-Union (mIoU), overall accuracy (oAcc) and mean accuracy (mAcc) across all classes.

\subsection{3D Unsupervised Semantic Segmentation}

\textbf{Evaluation on ScanNet-v2:} We conduct the unsupervised point cloud test on the ScanNet-v2 validation set and report the results in Tab.~\ref{tab:scannet}. As illustrated in Tab.~\ref{tab:scannet}, We compare our approach with prior unsupervised 3D point cloud segmentation methods, including PC-HC~\cite{hou2021contrastive}, PointDC~\cite{chen2023pointdc}, U3DS$^3$~\cite{liu2023u3ds3}, and LogoSP~\cite{zhang2025logosp}, as well as other baselines commonly reported in the literature, such as PiCIE~\cite{cho2021picie} and WYPR~\cite{ren2021wypr}. In the absence of any human annotations or pre-training on point cloud data, our method outperforms the majority of these baselines. Relative to the state-of-the-art LogoSP, we achieve a 0.9\% improvement in mIoU.

\begin{table}[htbp]
    \centering
    \caption{Comparison of unsupervised segmentation on the ScanNet-v2 validation set.}
    \small

        \begin{tabular}{l|c}
            \hline
            Unsupervised Methods & mIoU(\%) \\
            \hline\hline
            PC-HC~\cite{hou2021contrastive} & 4.6 \\
            PiCIE~\cite{cho2021picie} & 7.6 \\
            GrowSP~\cite{zhang2023growsp} & 25.4 \\
            PointDC~\cite{chen2023pointdc} & 25.7 \\
            U3DS³~\cite{liu2023u3ds3} & 27.3 \\
            WYPR~\cite{ren2021wypr} & 29.6 \\
            LogoSP~\cite{zhang2025logosp} & 35.8 \\
            \textbf{PointGS (Ours)} & \textbf{36.7} \\
            \hline
        \end{tabular}
    \label{tab:scannet}
\end{table}

\begin{table}[htbp]
    \centering
    \caption{Comparison of unsupervised segmentation on the S3DIS validation set (Area 5).}
    \small

        \begin{tabular}{l|c|c|c}
            \hline
            \thead{Unsupervised Methods} & mIoU(\%) & oAcc(\%) & mAcc(\%) \\
            \hline\hline
            PC-HC~\cite{hou2021contrastive} & 9.3 &  26.9 & - \\
            PiCIE~\cite{cho2021picie} & 17.8 & 46.4 & 28.1 \\
            WYPR~\cite{ren2021wypr} & 22.3 & - & - \\
            PointDC~\cite{chen2023pointdc} & 22.6 & 54.1 & - \\
            U3DS³~\cite{liu2023u3ds3} & 42.8 & 75.5 & 55.8\\
            GrowSP~\cite{zhang2023growsp} & 44.6 & 78.5 & 59.4 \\
            LogoSP~\cite{zhang2025logosp} & 46.5 & \textbf{82.8} & 55.9 \\
            \textbf{PointGS (Ours)} & \textbf{49.3} & 76.6 & \textbf{66.1} \\
            \hline
        \end{tabular}
    \label{tab:s3dis}
\end{table}

\medskip
\noindent\textbf{Evaluation on S3DIS:} We also evaluate the proposed method on the S3DIS Area 5 dataset to validate the effectiveness of the proposed method. Tab. ~\ref{tab:s3dis} presents a comparison with clustering-based unsupervised 3D point cloud segmentation methods, including U3DS³~\cite{liu2023u3ds3}, GrowSP~\cite{zhang2023growsp}, and LogoSP~\cite{zhang2025logosp}. As shown in the table, our approach surpasses the best baseline, achieving +2.8 mIoU improvement. The lower oAcc compared with GrowSP and LogoSP is mainly because oAcc is significantly affected by categories that contain a large number of points (such as \textit{ceiling}, \textit{wall} and \textit{floor}). Additionally, qualitative experiments conducted on the S3DIS Area 5 dataset, as depicted in Fig. \ref{fig:experiment}, provide further insights. The figure compares our method against state-of-the-art unsupervised approaches, including PointDC and GrowSP, demonstrating that our method more accurately localizes small objects. Moreover, our RGB-based point cloud projection method effectively segments near-planar objects, such as wall-mounted boards, as indicated by the red circles in Fig. \ref{fig:experiment}. Furthermore, our approach can identify and segment certain objects that extend beyond the annotated categories in the ground truth, illustrated by the green circles in Fig. \ref{fig:experiment}.

\begin{table}[htbp]
\centering
\caption{Impact of varying the number of projection views (\( V \)) on S3DIS Area 5.}
\small
\begin{tabular}{c|c}
\hline
\( V \) & S3DIS (mIoU\%) \\
\hline\hline
50 & 35.9 \\
75 & 42.2 \\
100 & 46.6 \\
125 & 48.9 \\
150 & 49.3 \\
200 & \textbf{49.4} \\
\hline
\end{tabular}
\label{tab:projection_views}
\end{table}

\begin{table}[htbp]
\centering
\caption{Impact of varying angular intervals on S3DIS Area 5.}
\small
\begin{tabular}{c|c|c}
\hline
\( \Delta_{\text{elev}} \) (\( \circ \)) & \( \Delta_{\text{azim}} \) (\( \circ \)) & S3DIS (mIoU\%) \\
\hline\hline
0.1 & 5.5 & 48.6 \\
0.3 & 6.5 & 49.1 \\
0.5 & 7.5 & \textbf{49.3} \\
0.7 & 8.5 & 47.3 \\
0.9 & 9.5 & 36.2 \\
\hline
\end{tabular}
\label{tab:angular_intervals}
\end{table}

\subsection{Ablation Experiment}
To showcase the effectiveness of each module, we conduct four groups of experiments on the S3DIS\cite{armeni2017s3dis} Area 5 dataset: (1) the baseline projection approach proposed without points to 3D Gaussians reconstruction in Sec. 3.3, (2) adding 3D-GS to the baseline without alignment. Since the Gaussian centers and the raw point cloud are not yet aligned in a unified coordinate system, this setting yields very poor performance. (3) Adding 2-Step ICP on the basis of the control group (2). The significantly enhanced performance indicates the necessity of 2-Step ICP for the 3D-GS method. (4) Adding Affinity Feature instead of directly aligning the rendered pixels with only the mask based on the control group (3), and (5) is our full model. As shown in Tab. ~\ref{tab:ablation}, our full model clearly outperforms the baseline on all of the evaluation metrics, benefiting from the 2-Step ICP, Affinity Feature, and Multi-View Consistency Check.

\begin{table}[htbp]
\begin{center}
\small
\caption{Ablation experiments of PointGS on the S3DIS Area5.}
\label{tab:ablation}
\begin{tabular}{cccc|c}
\hline
 \thead{3D-GS} & \thead{2-Step ICP} & \thead{Affinity \\ Feature} & \thead{Multi-View \\ Consistency Check} & \thead{mIoU\\(\%)} \\
\hline\hline
   & & & & 13.1\\
  $\checkmark$ & & & & 3.3 \\
  $\checkmark$ & $\checkmark$ & & & 27.5 \\
  $\checkmark$ & $\checkmark$ & $\checkmark$ & & 49.2 \\
  $\checkmark$ & $\checkmark$ & $\checkmark$ & $\checkmark$ & \textbf{49.3} \\
\hline
\end{tabular}
\end{center}
\end{table}

\subsection{Parameter Sensitivity Experiment}
We perform parameter sensitivity analysis on key projection hyperparameters---the number of views (\( V \)), angular intervals in elevation (\( \Delta_{\text{elev}} \)) and azimuth (\( \Delta_{\text{azim}} \)), and distribution type (surround or tiled)---evaluated on S3DIS Area 5 using unsupervised mIoU. As shown in Tab. ~\ref{tab:projection_views}, mIoU rises from 35.9\% at \( V = 50 \) to 49.3\% at \( V = 150 \), capturing richer geometric details for small/occluded objects, but plateaus at 49.4\% for \( V = 200 \); we select \( V = 150 \) to balance performance gains and resource costs. Table ~\ref{tab:angular_intervals} reveals peak mIoU (49.3\%) at \( \Delta_{\text{elev}} = 0.5\circ \), \( \Delta_{\text{azim}} = 7.5\circ \), with smaller angles risking incomplete coverage of scenes via more overlaps and larger ones weakening inter-frame correlations. Tab. ~\ref{tab:projection_distribution} favors surround (49.3\% mIoU) over tiled (45.9\%), as the circular path ensures uniform scene encapsulation, reducing blind spots in indoor scenes.

\begin{table}[htbp]
\centering
\caption{Comparison of projection distribution types on S3DIS Area 5.}
\small
\begin{tabular}{l|c}
\hline
Projection Distribution & S3DIS (mIoU\%) \\
\hline\hline
Surround & \textbf{49.3} \\
Tiled & 45.9 \\
\hline
\end{tabular}
\label{tab:projection_distribution}
\end{table}

\begin{table}[htbp]
\centering
\caption{Impact of varying cluster\_selection\_epsilon (\( \epsilon \)) with fixed min\_cluster\_size = 10 on S3DIS Area 5.}
\small
\begin{tabular}{c|c}
\hline
cluster\_selection\_epsilon & S3DIS (mIoU\%) \\
\hline\hline
0.05 & 46.9 \\
0.01 & \textbf{49.3} \\
0.005 & 49.1 \\
0.001 & 48.7 \\
\hline
\end{tabular}
\label{tab:sam_epsilon}
\end{table}

\begin{table}[htbp]
\centering
\caption{Impact of varying min\_cluster\_size (\( m \)) with fixed cluster\_selection\_epsilon = 0.01 on S3DIS Area 5.}
\small
\begin{tabular}{c|c}
\hline
min\_cluster\_size & S3DIS (mIoU\%) \\
\hline\hline
20 & 39.4 \\
15 & 44.5 \\
10 & \textbf{49.3} \\
5 & 49.4 \\
\hline
\end{tabular}
\label{tab:sam_min_size}
\end{table}

\begin{table}[htbp]
\centering
\caption{The impact of different segmentation granularity on S3DIS Area 5.}
\label{tab:scale_gate}
\small
\begin{tabular}{c|c}
\hline
Scale Gate & S3DIS (mIoU\%) \\
\hline\hline
0.2 & 46.6 \\
0.3 & 48.5 \\
0.4 & \textbf{49.3} \\
0.5 & 47.7 \\
0.6 & 35.1 \\
\hline
\end{tabular}
\end{table}

We further analyze SAM-specific parameters: \texttt{cluster\_selection\_epsilon} (\( \epsilon \)) for boundary sensitivity and \texttt{min\_cluster\_size} (\( m \)) for noise filtering, to prevent over/under-segmentation. Table ~\ref{tab:sam_epsilon} (fixed \( m = 10 \)) shows mIoU increasing from 46.9\% at \( \epsilon = 0.05 \) to 49.3\% at \( \epsilon = 0.01 \), enhancing edge precision, but diminishing to 48.7\% at 0.001 due to noise sensitivity; we choose \( \epsilon = 0.01 \) for balanced clustering. Table ~\ref{tab:sam_min_size} (fixed \( \epsilon = 0.01 \)) indicates mIoU rising from 39.4\% at \( m = 20 \) to 49.3\% at \( m = 10 \), better capturing compact objects, with marginal gain (49.4\%) at \( m = 5 \) risking noise; we select \( m = 10 \) to optimize small-structure detection.

We also test the impact of the segmentation granularity size on the segmentation performance for different value of Scale Gate $s$ in Table ~\ref{tab:scale_gate}. It can be seen that the segmentation performance reaches its optimum when the Scale Gate is 0.4. A smaller Scale Gate value will amplify the channels in the features corresponding to fine-grained segmentation (such as object components). Although it can improve the segmentation accuracy of smaller items, it sacrifices the semantic consistency of larger items. While a larger Scale Gate value will suppress these channels, thereby highlighting coarse-grained targets (such as the entire object), it results in poor recognition ability for smaller objects. This parameter needs to be adjusted according to different scenes. For a more detailed and complex ScanNet scene involving more small items, the performance is the best when the Scale Gate is set to 0.3.

\section{Conclusion}

We present a novel framework for unsupervised semantic segmentation of 3D point clouds, requiring no human annotations or pre-training on point cloud data, named PointGS. The process begins by reconstructing the point cloud into 3D Gaussians from multi-view images, leveraging 3D Gaussian Splatting to maintain spatial information and resolve the foreground-background overlap issues caused by point sparsity. Concurrently, SAM-generated 2D masks from these views are used in contrastive learning with the rendered Gaussians to derive semantically segmented Gaussians, ensuring cross-view consistency through multi-view depth checks and semantic uniqueness. Subsequently, the segmented Gaussians are aligned with the original point cloud via a spatial process that extracts center kernel points, computes ratios, performs two-step registration to avoid local optima, and propagates labels through nearest-neighbor assignment. Evaluation on benchmarks demonstrates superior performance, with gains of +2.8\% mIoU on S3DIS and +0.9\% mIoU on ScanNet-v2.
\section{Acknowledgments}

This work was supported in part by the Fundamental Research Funds for the Central Universities under Grant 2024QYBS026; in part by Beijing Natural Science Foundation under Grant L231019; in part by the National Natural Science Foundation of China under Grant 62276019, 62306028, 62501043 and U22B2004; in part by Langfang Research and Development Projects under Grant 2023011003B; and in part by Shenzhen Science and Technology Program Project under Grant KJZD20240903102742055; Open Grants of Key Laboratory of Lightning, China Meteorological Administration (No. 2024KELL-A001).

\clearpage
{
    \small
    \bibliographystyle{ieeenat_fullname}
    \bibliography{main}
}
\clearpage
\setcounter{page}{1}
\maketitlesupplementary

\begin{strip}
  \centering
  \includegraphics[width=\textwidth]{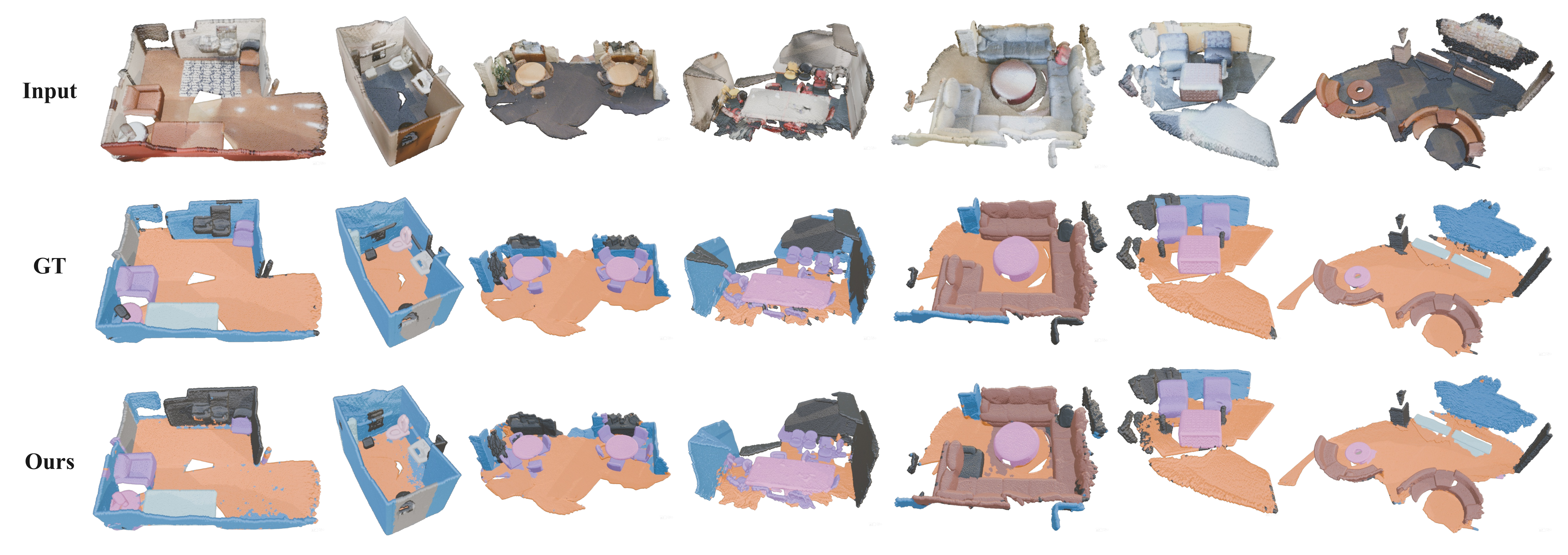}
  \captionof{figure}{Qualitative comparison with Ground Truth of unsupervised segmentation on the ScanNet validation set. Each color represents one semantic class.}
  \label{fig:scannet}
\end{strip}

\section{Additional Experiments}
\label{sec:rationale}

Due to space constraints in the main text, some of the experiments are placed in this supplementary material.


\subsection{Qualitative Experiment on ScanNet-v2}
To verify the performance on the ScanNet-v2 dataset, we conduct additional visual experiments on this dataset. As shown in Fig. \ref{fig:scannet}, the performance of our method on the ScanNet dataset is very close to the Ground Truth. For more open rooms, we directly projected the scene; for more enclosed rooms, we first split the scene into two parts and then performed projection, following the strategy used for S3DIS.

\subsection{Additional Contrast Experiment}
\begin{figure*}[t]
  \centering
  \includegraphics[width=1\linewidth]{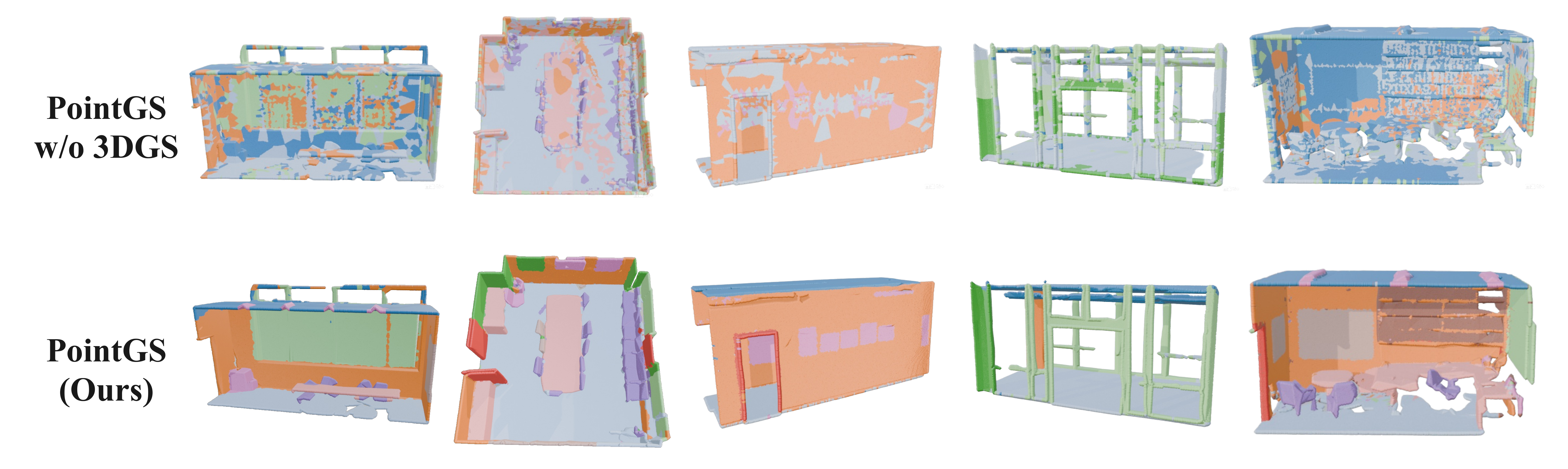}
  \caption{
     Qualitative comparison of different ablation settings on the ScanNet validation set. 
  }
  \label{fig:sam}
\end{figure*}

We present qualitative results of the direct point projection method without 3D-GS as a comparison to our approach. The qualitative experimental results are presented in Fig. \ref{fig:sam}. As can be seen from Fig. \ref{fig:sam}, although the general semantic boundaries can be identified, the semantic ambiguity occurs due to the confusion between foreground and background. Since the semantics were directly grown from the initial sparse point cloud onto the original point cloud, the areas of semantic confusion present as large patches.


\subsection{Additional Experiment on S3DIS}
 We also test the performance of S3DIS Area 5 in terms of mIoU for each category. The per-category mIoU from label 0 to 12 are respectively 68.4/ 74.5/ 57.5/ 40.2/ 51.6/ 44.9/ 43.6/ 39.0/ 50.1/ 65.5/ 38.3/ 45.3/ 31.5. The mIoU of S3DIS 6-fold validation is 45.8.

\end{document}